\documentclass[final,runningheads]{llncs}
\usepackage[english]{babel}
\usepackage{graphicx}
\usepackage{tabularx}
\usepackage{amsmath}
\usepackage{hyperref}
\usepackage{cleveref}
\Crefformat{figure}{#2Fig.~#1#3}
\Crefformat{section}{#2Sect.~#1#3}

\usepackage{paralist}

\usepackage{caption} 

\usepackage{listings}

\usepackage{caption}
\usepackage{subcaption}

\usepackage[obeyFinal,textsize=scriptsize,backgroundcolor=yellow!40]{todonotes}

\usepackage[nonumberlist,acronym,sanitize=none]{glossaries}

\usepackage{color}
\usepackage[T1]{fontenc}

\makeatletter
\lst@InstallKeywords k{aftercolon}{aftercolonstyle}\slshape{aftercolonstyle}{}ld
\makeatother

\makeatletter
\lst@InstallKeywords k{prefixes}{prefixestyle}\slshape{prefixestyle}{}ld
\makeatother

\makeatletter
\lst@InstallKeywords k{variables}{variablestyle}\slshape{variablestyle}{}ld
\makeatother

\graphicspath{{imgs/}}

{\begin{inparaenum}[\itshape(i)\upshape]}%
{\end{inparaenum}}

\newcommand{\resultsTableFontSize}{\scriptsize}
\usepackage{todonotes}

\begin{document}
\counterwithin{lstlisting}{section}
\title{Describing and Organizing Semantic Web and Machine Learning Systems in the SWeMLS-KG}
\titlerunning{Semantic Web and Machine Learning Systems Knowledge Graphs}
\author{
Fajar J. Ekaputra\inst{1,2}
\and
Majlinda Llugiqi\inst{1}
\and
Marta Sabou\inst{1}
\and
Andreas Ekelhart\inst{3,4}
\and
Heiko Paulheim\inst{5}
\and
Anna Breit\inst{6}
\and
Artem Revenko\inst{6}
\and
Laura Waltersdorfer\inst{2}
\and \\
Kheir Eddine Farfar\inst{7}
\and
Sören Auer\inst{7,8}
}

\authorrunning{F.J. Ekaputra et al.}

\institute{
WU (Vienna University of Economics and Business) 
\email{first.last@wu.ac.at}
\and
TU Wien
\email{first.last@tuwien.ac.at}
\and
University of Vienna 
\email{first.last@univie.ac.at}
\and
SBA Research 
\email{first.last@sba-research.org}
\and
University of Mannheim 
\email{first.last@uni-mannheim.de}
\and
Semantic Web Company 
\email{first.last@semantic-web.com}
\and
TIB Leibniz Information Centre for Science and Society
\email{first.last@tib.eu}
\and
L3S Research Center, Leibniz University of Hannover
\email{auer@l3s.de}
}

\maketitle              
\vspace{-5mm}
\begin{abstract}
The overall AI trend of creating neuro-symbolic systems is reflected in the Semantic Web community with an increased interest in the development of systems that rely on both \textit{Semantic Web resources} and \textit{Machine Learning components} (SWeMLS, for short).
However, understanding trends and best practices in this rapidly growing field is hampered by a lack of standardized descriptions of these systems and an annotated corpus of such systems. 
To address these gaps, we leverage the results of a large-scale systematic mapping study collecting information about 470 SWeMLS papers and formalize these into one resource containing: (i) the \textit{SWeMLS ontology}, (ii) the \textit{SWeMLS pattern library} containing machine-actionable descriptions of 45 frequently occurring SWeMLS workflows, and (iii) \textit{SWEMLS-KG}, a knowledge graph including machine-actionable metadata of the papers in terms of the SWeMLS ontology. This resource provides the first framework for semantically describing and organizing SWeMLS thus making a key impact in (1) understanding the status quo of the field based on the published paper corpus and (2) enticing the uptake of machine-processable system documentation in the SWeMLS area.  

\end{abstract}

\noindent \textbf{Resource type:} Knowledge Graph

\noindent \textbf{License:} \url{https://creativecommons.org/licenses/by/4.0/}

\noindent \textbf{DOI:} \url{https://doi.org/10.5281/zenodo.7445917}

\noindent \textbf{URL:} \url{https://w3id.org/semsys/sites/swemls-kg/}

\section{Introduction}
\label{sec:intro}

The field of Artificial Intelligence (AI) is currently witnessing a great interest in (more closely) integrating and bridging between symbolic and sub-symbolic (AI)~\cite{booch2020thinking} techniques. This substantial trend led to the establishment of the new sub-research field of \textit{neuro-symbolic systems}\footnote{The Neurosymbolic Artificial Intelligence journal will be launched in 2023: \url{https://www.iospress.com/catalog/journals/neurosymbolic-artificial-intelligence}}~\cite{besold2021neural,garcez2002neural}, which focuses on the theoretical and practical aspects of creating such complex systems. Against this backdrop, it is not surprising that this AI trend is also reflected in  the Semantic Web (SW) research community which has popularized AI-based knowledge representation techniques and resources in the last two decades~\cite{hitzler2021review}. There is increased interest in neuro-symbolic integration in the context of the Semantic Web~\cite{HitzlerNeSy2020}, such as the development of systems that rely on both Semantic Web resources and Machine Learning components. We coined the term \textit{Semantic Web and Machine Learning System} (SWeMLS) to refer to such systems \cite{breit2022combining}.

\begin{figure}[t]
 \centering
    \includegraphics[width=0.7\columnwidth]{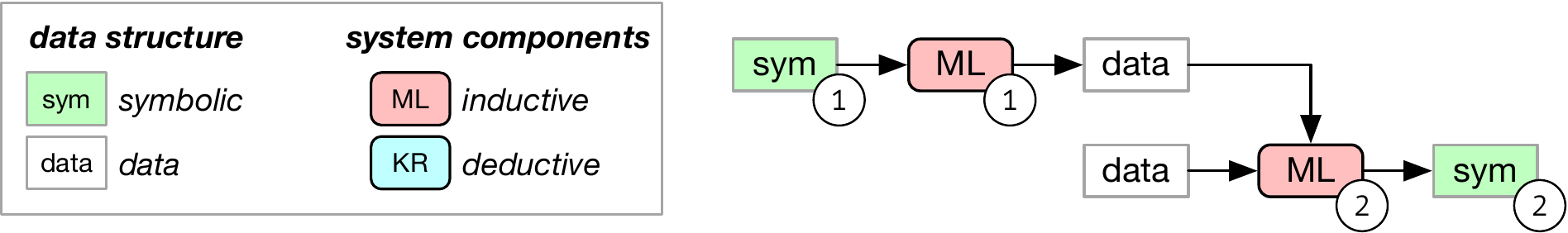}
    \caption{Schematic representation of a SWeMLS workflow for art classification~\cite{Garcia2019}.}
    \label{fig:pattern-t3}
    \vspace{-5mm}
\end{figure}

For example, in~\cite{Garcia2019} authors propose a system for automatic art analysis that can be classified as an SWeMLS. To that end, they augment a deep learning based system that classifies artistic images purely based on visual features with contextual art information in a form of a knowledge graph about  painters, paintings,  artistic schools, etc. Schematically, the system's workflow is depicted in Fig.~\ref{fig:pattern-t3} with the boxology notation introduced by~\cite{VanHarmelen2019a}: starting with the \texttt{sym1} knowledge graph, graph embeddings are created through \texttt{ML1} which is a CNN deep learning model (\texttt{sym1-ML1-data}); subsequently, these embeddings together with visual data (i.e., images) are input to a CNN model (\texttt{ML2}) to create image classifications (\texttt{sym2}). Authors experimentally show that the inclusion of the SW component leads to performance increases by 7.3\% in art classification and 37.24\% in image retrieval tasks, thus demonstrating the potential of such systems.

Given the potential of SWeMLS and the increased interest in this field, the key \textit{motivation} for our work was to gain a systematic understanding of the SWeMLS area by identifying trends among such systems and clustering them to better characterize the landscape of published systems. The main \textit{challenges} in achieving a large-scale, data-driven, representative and systematic analysis of the SWeMLS field were:
\begin{itemize}
\item (i) a lack of understanding of important \textit{system characteristics} that should be considered when analyzing SWeMLS. Approaches to characterise neuro-symbolic systems either focus on broader families of systems than SWeMLS~\cite{bader2005dimensions},\\~\cite{kautzthird}, or on a specific aspect of the systems (e.g., their internal processing flow~\cite{VanHarmelen2019a,VanBekkum2021}). Additionally, none are formalized for the purpose of using them as a basis for machine-actionable descriptions of the systems.
\item (ii) the lack of a corpus of systematically collected (and therefore representative) papers annotated in terms of such characteristics, to allow for a data-driven research trend analysis.  While a number of papers about systems that learn and reason were collected as a basis for the analysis described in~\cite{VanHarmelen2019a}, these were not offered as a corpus of annotated papers to the community.  
\end{itemize}

We addressed both challenges by conducting a large-scale Systematic Mapping Study (SMS~\cite{kitchenham2007guidelines}) on SWeMLS~\cite{breit2022combining}, through which we (i) proposed a set of characteristics for describing SWeMLS and (ii) systematically collected, selected and extracted data from nearly 500 papers describing such systems. This led to the following artifacts which together are offered as one \textit{resource}:
\begin{itemize}
    \item \textit{the SWeMLS ontology} that describes the main aspects of SWeMLS including their internal workflow in terms of boxology patterns as shown in Fig.~\ref{fig:pattern-t3}. The ontology schema (i.e., capturing important SWeMLS characteristics, e.g., $StatisticalModel$) and relevant instances (e.g., $DeepLearningModel$) were derived systematically during the scoping and analysis phases of the SMS,

    \item \textit{the SWeMLS-KG}: a knowledge graph containing the machine-actionable description of almost 500 systems in terms of the SWeMLS ontology, and

    \item \textit{the SWeMLS Pattern Library} containing the machine-actionable description of 45 SWeMLS patterns and their associated SHACL-based validation constraints. This pattern library extends the initial pattern catalog of 10-15 patterns originally identified by~\cite{VanHarmelen2019a} both \textit{quantitatively} with additional patterns observed during the SMS and \textit{qualitatively}, by offering the patterns in a machine-actionable rather than graphical representation. 
\end{itemize}

This resource is \textit{timely} considering the recent trend in the SW community (and beyond) to create systems that leverage both SW and ML components. To the best of our knowledge, it is also \textit{novel} by (i) providing the first ontology (and associated pattern library) for describing SWeMLS in a machine-actionable way and (ii) a methodologically collected corpus of SWeMLS and their semantic description. The resource is of immediate benefit for (SW) researchers that aim to explore trends in the SWeMLS field by analysing the data in the SWeMLS-KG and as such promises to have an impact on the understanding of the status-quo in this emerging field. Furthermore, the resource provides a semantic framework for describing SWeMLS and their internal details, thus potentially strongly influencing this field in terms of being well-documented, data-driven, and transparent.  

We  continue by discussing the impact of this resource (\Cref{sec:impact}) and then detail its main components and the methodology used to produce them (\Cref{sec:quality}), availability (\Cref{sec:availability}) and usage in two use cases (\Cref{sec:reusability}).
We summarize related work in \Cref{sec:relatedwork} and conclude with an outlook on future work in \Cref{sec:conclusion}.

\section{Impact of the Resource}
\label{sec:impact}

This resource is interesting to the Semantic Web community both in terms of its immediate and potential future impact on the field. An \textit{immediate impact} is \textit{enabling the understanding of general trends in the emerging area of SWeMLS}. The SWeMLS-KG allows for the first time to perform data-driven analysis in order to better understand  this family of systems. This can be achieved as part of two scenarios as described next and in more detail in Sections~\ref{sec:querying} and~\ref{sec:embedding}.
\begin{itemize}
    \item \textit{Asking concrete research questions},  e.g., \textit{What kind of processing patterns are the most frequent? Which ML methods are used most often in combination with which SW resources?} Such targeted analysis was performed as part of the SMS~\cite{breit2022combining} from which the SWeMLS ontology and KG were derived. While we investigated a limited number of questions that were feasible within the scope of the SMS, by making this resource available openly we enable the research community at large to perform additional analysis. 
    \item \textit{Identifying new insights} (e.g., through graph embedding) 
    allows uncovering a new understanding of the field by exploring latent semantics encoded in the data. 
\end{itemize}

Furthermore, the presented resources could have an important future impact by enabling the following use cases:

\begin{itemize}
\item \textit{Search for SWeMLS-related work.} Researchers that create a SWeMLS could more easily find related systems, as part of related work search, during the design, evaluation, and publishing of their own systems. The current resource supports answering questions such as: \textit{Which system patterns/pattern types are most frequent for graph completion tasks in the medical domain?}

\item \textit{Machine readable documentation and validation of SWeMLS.} Researchers that want to document a SWeMLS, can now (1) describe the system in a machine-processable way in terms of the SWeMLS ontology and (2) verify the correctness of their description through the SHACL validation. While the core technical artifacts are in place to enable other researchers to document their systems, future work will focus on more user-friendly annotation tools to entice large-scale adoption of research documentation for SWeMLS. 

\item \textit{Improved scientific reviewing and publication processes.} AI-related conferences are struggling with high numbers of submissions which leads to challenges (i) for conference organisers to meaningfully assign papers to reviewers; as well as (ii) for reviewers who are overloaded with receiving very diverse papers and challenged to compare new systems to related work. We envision that SWeMLS-related events could use the SWeMLS ontology as a basis  for annotating the submitted system papers. Such in-depth annotation of the systems could support (i) assigning relevant/similar papers to reviewers by  clustering papers in terms of (the intersection of) several dimensions (task solved, domain addressed, system pattern used); (ii) allow reviewers to more easily comprehend the design of the system by referring to a structured or even visual notation of the system besides its textual description in the paper. Naturally, reviewers could leverage other collections of annotated systems (e.g., the SWeMLS KG), to identify papers similar to the one reviewed to make an informed assessment of novelty. 
\end{itemize}

To conclude, the proposed resource could have a major impact on the way the research-documentation-publication cycles of SWeMLS happen, leading to a data-driven field and supporting faster growth and shorter innovation cycles.

\section{Knowledge Graph Construction}
\label{sec:quality}

We hereby describe the overall methodology for the construction of SWeMLS-KG (Sect.~\ref{sec:kg-creation}) and focus on key elements in this process such as the  SWeMLS ontology (Sect.~\ref{sec:ontology}), the SWeMLS pattern library (Sect.~\ref{sec:patternLibrary}) and the SWeMLS-KG and its population process (Sect.~\ref{sec:populationupdate}).

\begin{figure}[t]
 \centering
    \includegraphics[width=0.9\columnwidth]{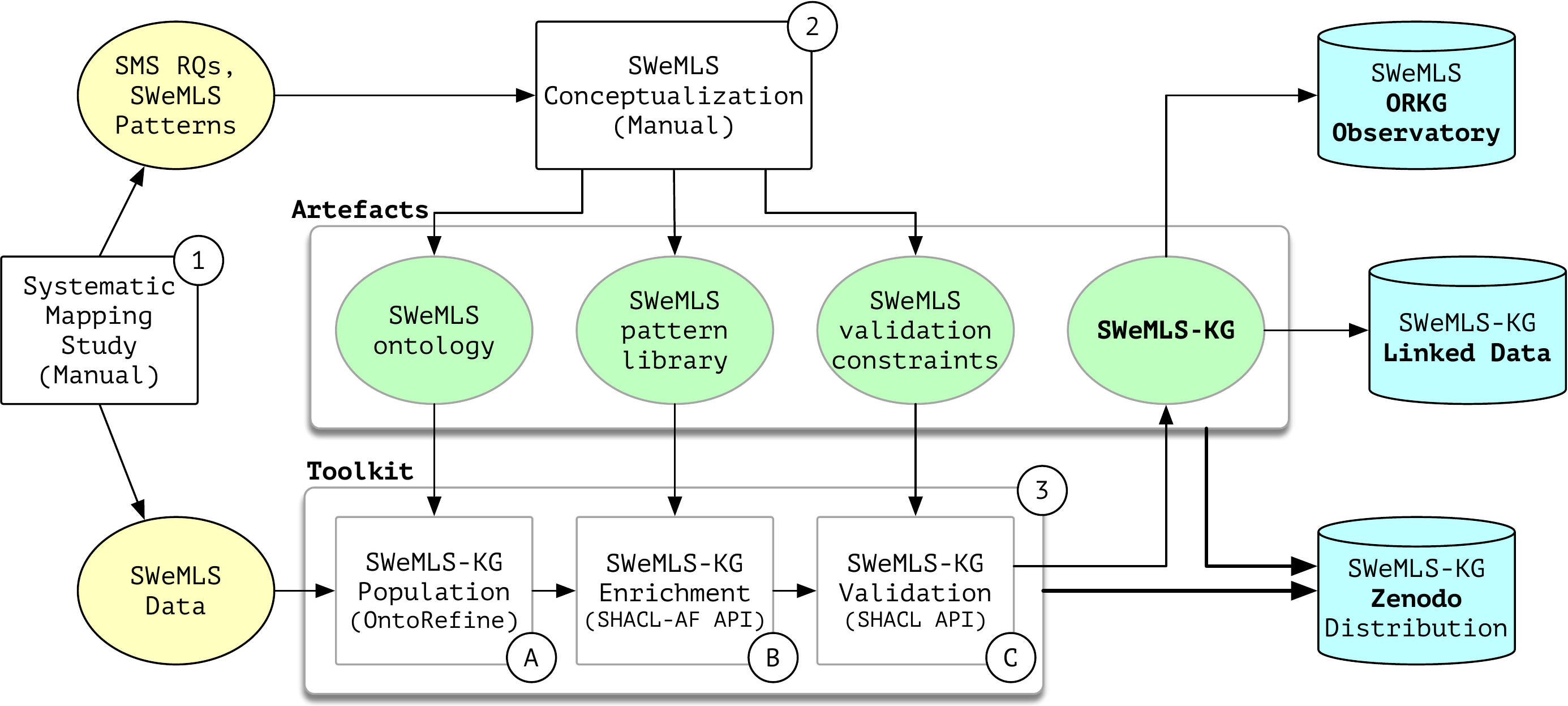}
    \caption{Overview of the SWeMLS-KG construction process}
    \label{fig:swemls-process}
    \vspace{-5mm}
\end{figure}
\subsection{SWeMLS-KG Construction Process and Methodology}
\label{sec:kg-creation}
Fig.~\ref{fig:swemls-process} depicts the process and methodology followed to construct the SWeMLS-KG. The starting point for the process was our prior large-scale SMS on the topic of SWeMLS~\cite{breit2022combining} (cf. \textbf{Step 1} in \Cref{fig:swemls-process}) during which we collected information from 476 papers on SWeMLS in spreadsheet format. Starting from this SMS, we converted its results into a machine-processable format, through the next steps. 

In \textbf{Step 2} we \textit{conceptualised the SWeMLS related information} from two inputs provided in the SMS results: (i) the SMS research questions were a basis for the competency questions of the SWeMLS ontology, and (ii) the 45 SWeMLS patterns identified from the papers, which have been described and depicted as drawings but were not yet formalized in a machine-processable manner.
From this step, we produce three types of outputs: (a) the SWeMLS ontology (\Cref{sec:ontology}), (b) the SWeMLS pattern library, which consists of pattern templates represented as RDF instances and SHACL-Advanced Features (SHACL-AF) rules, and (c) SWeMLS constraint definitions, which provide users with a mean to validate SWeMLS instances based on the existing patterns.

In \textbf{Step 3}  we perform the \textit{population of SWeMLS-KG} using the SWeMLS data and the artifacts produced in Step 2. This step (detailed in \Cref{sec:populationupdate}) consists of three sub-steps: (3A) populating the KG with spreadsheet data extracted from the SMS, (3B) constructing workflows between components and variables linked to each SWeMLS, and (3C) validating the KG using pattern-specific validations. The integrated and validated SWeMLS-KG is published through three distribution channels: (i) Linked Data interface, (ii) ORKG observatory, and (iii) a Zenodo repository as further explained in \Cref{sec:availability}.

\subsection{The SWeMLS Ontology}
\label{sec:ontology}

\begin{figure}[t]
 \centering
    \includegraphics[width=\columnwidth]{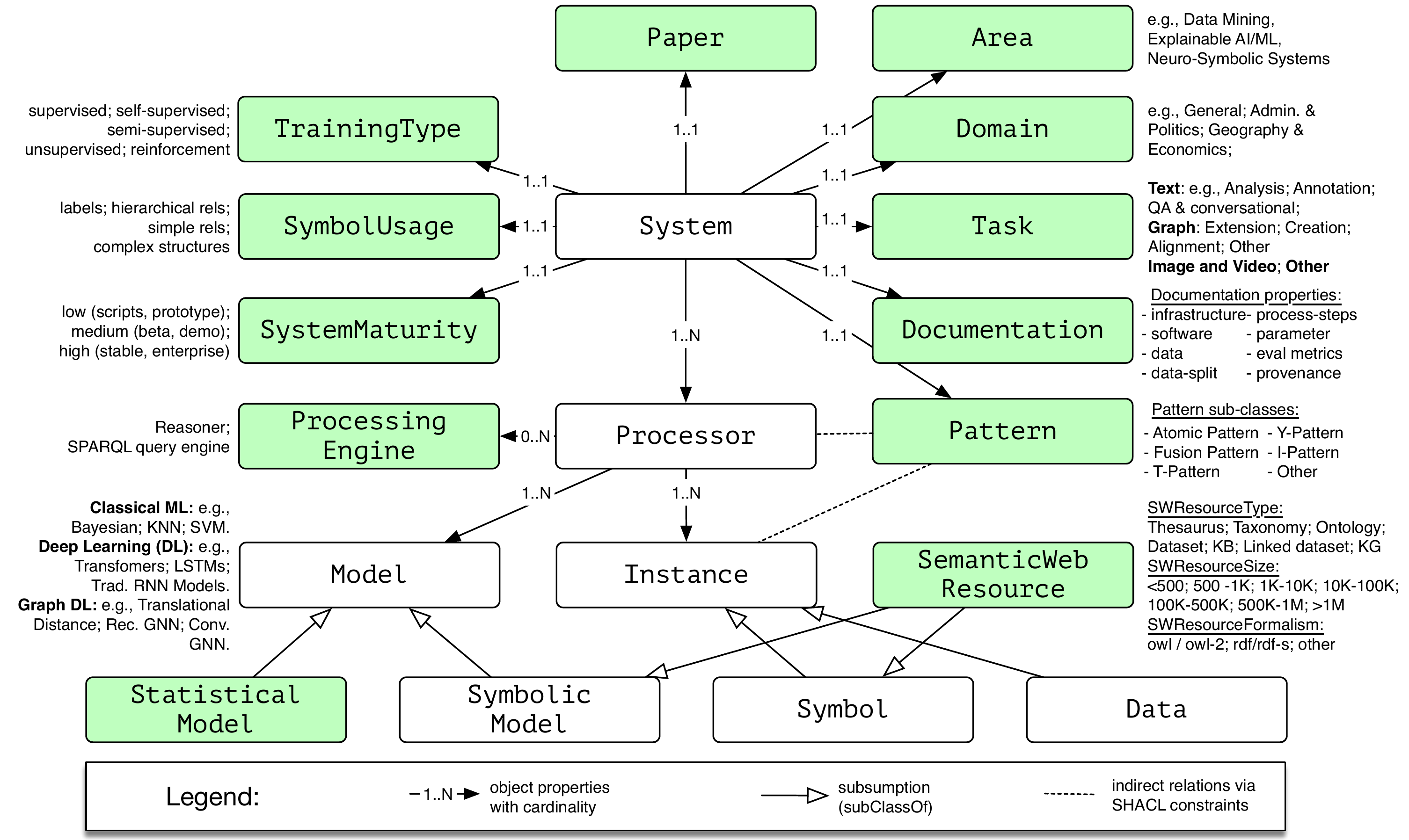}
    \caption{SWeMLS Ontology Overview (adapted from \cite{breit2022combining})}
    \label{figure:ontology_overview}
    \vspace{-5mm}
\end{figure}

\subsubsection{Ontology Creation}
\label{sec:ontology-creation}
To create the ontology, we followed the Ontology Development 101 guideline~\cite{noy2001ontology}. 
We started by determining the domain and the scope of the ontology, using SMS research questions as competency questions: 
\begin{itemize}
    \item \textbf{Bibliographic characteristics} - How are the publications temporally and geographically distributed? How are the systems positioned, and which keywords are used to describe them?
    \item \textbf{System Architecture} - What processing patterns are used in terms of inputs/outputs and what is the order of processing units?
    \item \textbf{Application Areas} - What kind of tasks are solved (e.g., text analysis)? In which domains are SWeMLS applied (e.g., life sciences)?
    \item \textbf{Characteristics of the ML Module} - What ML models are incorporated (e.g., SVM)? Which ML components can be identified (e.g., attention)? What training type(s) is used during the system training phase?
    \item \textbf{Characteristics of the SW Module} - What type of Semantic Web structure is used (e.g., taxonomy)? What is the degree of semantic exploitation? What are the size and the formalism of the resources? Does the system integrate semantic processing modules (i.e., KR)?
    \item \textbf{Maturity, Transparency and Auditability} - What is the level of maturity of the systems? How transparent are the systems in terms of sharing source code, details of infrastructure and evaluation setup? Does the system have a provenance-capturing mechanism? 
\end{itemize}

We considered reusing existing ontologies, especially to represent the patterns' workflows such as Wings Workflow~\cite{GilRKGGMD11}, Taverna~\cite{HullWSGPLO06}, and the Common Workflow Language~\cite{amstutz2016common}. However, our patterns are very specific and none of them could be used.
Thus, we decided to develop our own SWeMLS ontology by adapting and extending the P-PLAN ontology~\cite{GarijoGC14} to describe SWeMLS workflows and the OPMW ontology~\cite{MoreauCFFGGKMMMPSSB11} to describe system patterns.

As the next step, we enumerated important terms from the SMS that should be represented in the ontology such as \textit{System}, \textit{Paper}, \textit{Processor}, \textit{Model}, and \textit{Instance}. 
Then we defined the classes and the class hierarchy based on these terms, using a top-down approach, e.g., class \textit{Model} has \textit{Semantic Model} and \textit{Statistical Model} as sub-classes. 
After establishing the classes and their hierarchy, we defined the class properties based on the data gathered from the SMS, e.g., system application area, task, and system maturity.
Finally, we created individuals from the SMS data, i.e., \textit{Data Mining}  is an instance of \textit{Area}.

\subsubsection{Ontology Description}
The resulting SWeMLS ontology is intended to represent the systems described in the publications reported in~\cite{breit2022combining}. 
A high-level overview of its main classes, properties, and an excerpt of named individuals is shown in \Cref{figure:ontology_overview}. 
Overall, the SWeMLS ontology includes:
(i) paper details, (ii) system properties reported in such papers, and (iii) workflow-style representations of patterns:
\begin{itemize}
    \item \textbf{Paper} details such as title, year of \textit{publication}, \textit{publication type}, \textit{venue}, \textit{authors' countries}, \textit{keywords}, a short \textit{summary}, and the \textit{link} to the paper.
    \item \textbf{SWeMLS} properties such as the \textit{targeted tasks}, \textit{level of maturity}, \textit{application domain}, \textit{semantic web resources} being used, \textit{machine learning model}, type of \textit{semantic processor}, \textit{the pattern} being used, as well as \textit{documentation} properties which include: e.g., \textit{infrastructure}, \textit{provenance}, and \textit{evaluation}.
    \item \textbf{SWeMLS patterns} representing the structure of each system workflow pattern with each pattern's component including their inputs/outputs. We detail the representation of the SWeMLS patterns in Sect.~\ref{sec:patternLibrary}. 
\end{itemize}

An example of how the terms defined by the ontology are used to describe a paper reporting a SWeMLS is presented in Sect.~\ref{sec:kg-creation}.

\subsection{SWeMLS Pattern Library}\label{sec:patternLibrary}

\paragraph{Pattern representation.} We use the P-PLAN~\cite{GarijoGC14} and OPMW ~\cite{MoreauCFFGGKMMMPSSB11} as the basis for SWeMLS pattern representation. More specifically, we follow the separation of three major types of workflow structures outlined by Garijo et al.~\cite{GarijoGC14}: 
\begin{itemize}
    \item (i) Workflow Template (\texttt{opmw:WorkflowTemplate}), a generic pattern that indicates the type of steps in the workflow and their dataflow dependencies, 
    \item (ii) Workflow Instance (\texttt{swemls:System} as a sub-class of \texttt{p-plan:Plan}), a workflow that specifies the application algorithms to be executed and data to be used, and 
    \item (iii) Workflow Execution (\texttt{p-plan:Bundle}), a workflow execution trace containing details of what happened during an execution. 
\end{itemize}

We focus on the first two types (i.e., Workflow Template and Instance) and plan for Workflow Execution as part of our future work. 

\paragraph{The SWeMLS Pattern Library} consists of the representation of 45 system processing patterns identified during the SMS (each pattern is captured in a .ttl file in the zenodo distribution of the resource). An example representation of the Workflow Template for pattern T-3 is shown in Listing \ref{lst:t3pattern}. The template contains the definition of the patterns (e.g., \texttt{res:Pattern.T3}), its components/steps (i.e., \texttt{res:Pattern.T3.ML1} and \texttt{res:Pattern.T3.ML2}) and how they use or generate variables (e.g., \texttt{res:Pattern.T3.ML1} use \texttt{res:Pattern.T3.SW1} and generate \texttt{res:Pattern.T3.Data2}).\\

\vspace{-2mm}
\begin{lstlisting}[
captionpos=b, 
caption={T-3 pattern in turtle format}, 
label=lst:t3pattern,
basicstyle=\scriptsize\ttfamily]
@prefix swemls: <https://w3id.org/semsys/ns/swemls#> .
@prefix res: <http://semantic-systems.net/swemls/> .
@prefix p-plan: <http://purl.org/net/p-plan#> .
@prefix opmw: <http://www.opmw.org/ontology/> . 
/* ... rdf, rdfs are ommitted */

/* Pattern T3 as an instance of opmw:WorkflowTemplate */
res:Pattern.T3 a opmw:WorkflowTemplate ; rdfs:label "T3";
  rdfs:comment "[{sym -> ML -> data / data} -> ML -> sym]" .

/* Component T3.ML1 with T3.SW1 (SW resource) as input */
res:Pattern.T3.ML1 a swemls:WorkflowTemplateProcessML ;
  opmw:isStepOfTemplate res:Pattern.T3 ; opmw:uses res:Pattern.T3.SW1 .
/* Component T3.ML2 with T3.Data1 and T3.Data2 (data) as inputs */
res:Pattern.T3.ML2 a swemls:WorkflowTemplateProcessML ; 
  p-plan:isPreceededBy res:Pattern.T3.ML1 ; 
  opmw:isStepOfTemplate res:Pattern.T3; opmw:uses res:Pattern.T3.Data1, res:Pattern.T3.Data2 .

/* Variable T3.SW1 */
res:Pattern.T3.SW1 a swemls:TemplateArtifactSW ; opmw:isVariableOfTemplate res:Pattern.T3 .
/* Variable T3.Data1; used as input for component T3.ML2 */
res:Pattern.T3.Data1 a swemls:TemplateArtifactData; opmw:isVariableOfTemplate res:Pattern.T3 .
/* Variable T3.Data2; generated by T3.ML1; input for T3.ML2 */
res:Pattern.T3.Data2 a swemls:TemplateArtifactData ;
  opmw:isVariableOfTemplate res:Pattern.T3 ; opmw:isGeneratedBy res:Pattern.T3.ML1 .

/* Variable T3.SW2; generated by T3.ML2 as the final result of the System */
res:Pattern.T3.SW2 a swemls:TemplateArtifactSW ;
  opmw:isVariableOfTemplate res:Pattern.T3 ; opmw:isGeneratedBy res:Pattern.T3.ML2 .

\end{lstlisting}
\vspace{-5mm}
\begin{figure}[t]
 \centering
    \includegraphics[width=0.9\columnwidth]{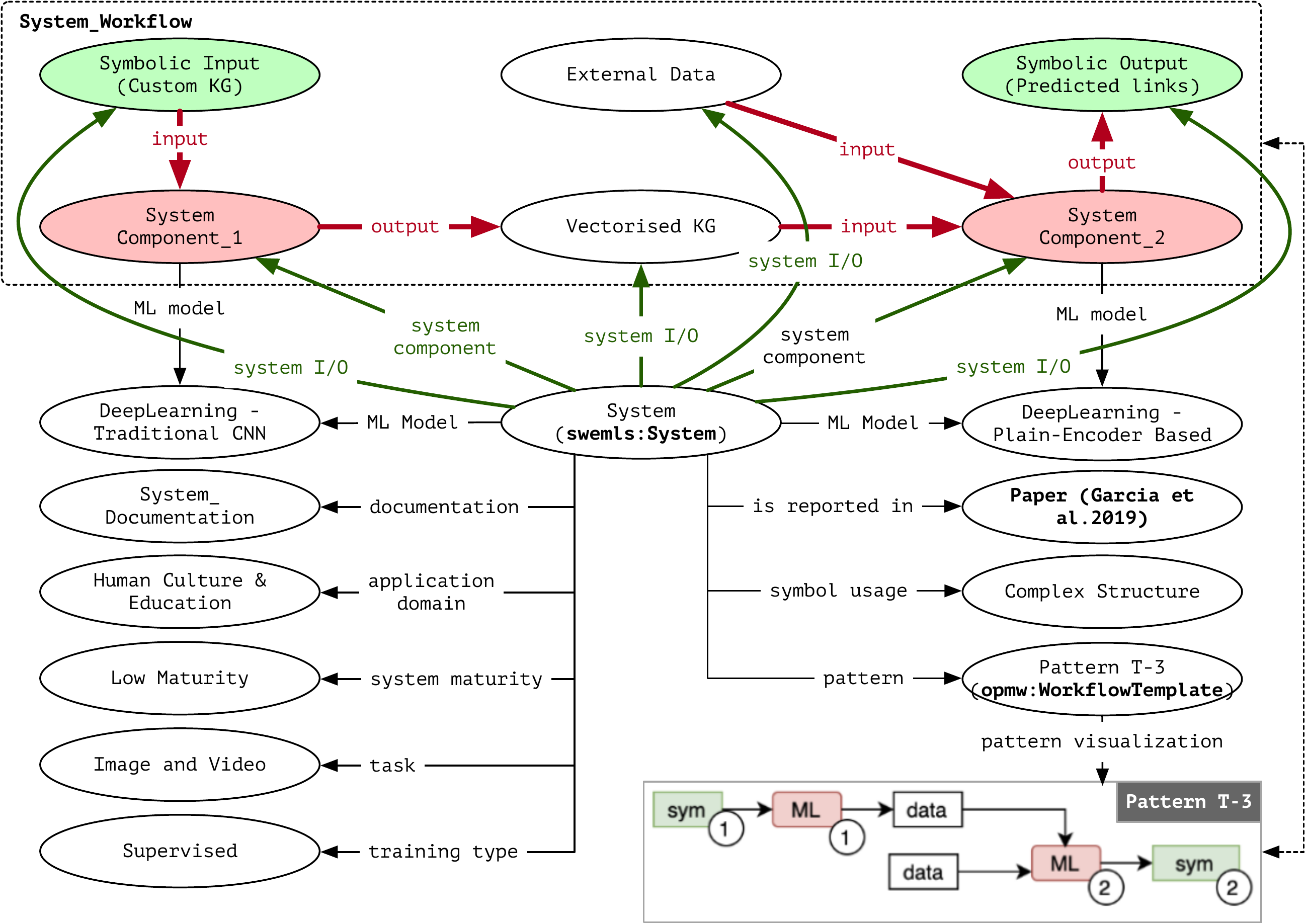}
    \caption{Example of the semantic representation of paper~\cite{Garcia2019} and the art classification SWeMLS described by it (adapted from \cite{breit2022combining}). \textit{Green arrows} represent relations between a system and its components and variables, \textit{Red arrows} represent workflow information generated with SHACL-AF rules.}
    \label{figure:example_instance_KG}
    \vspace{-7mm}
\end{figure}

\subsection{SWeML-KG Population and Update Mechanisms}
\label{sec:populationupdate}
After defining the underlying ontology, we populated the SWeMLS KG with the details of SWeMLS collected in the course of the SMS. 

\textit{An example SWeMLS semantic description} is depicted in \Cref{figure:example_instance_KG} which shows the SWeMLS-KG instance of the SWeMLS discussed in \Cref{sec:intro} \cite{Garcia2019}. For this system, we display the paper in which it is reported, together with other paper details, such as title, keywords, and year of publication. In addition, the target task to be solved falls into the category of '\textit{Image and Video}', the system maturity is reported as '\textit{Low}', the application domain is listed as '\textit{Human Culture and Education}', the training type as '\textit{Supervised}' and the symbol usage as '\textit{Complex Structure}'. 
The depicted system also contains documentation information, including transparency and auditability components of the system. 

The chosen system instantiates pattern ``T-3" (depicted visually in the bottom right of \Cref{figure:example_instance_KG}) which involves two symbolic data components, two machine learning components, and two data components in its workflow. The upper part of \Cref{figure:example_instance_KG} shows the semantic representation of the system workflow: starting from  custom KG as symbolic input to a traditional CNN machine-learning component,  the model produces a vectorized KG, which, along with some external data, serves as input for a deep learning plain encoder-based model. Finally, the system produces predicted links as symbolic output.

\textit{The SWeMLS population process} consisted of the following steps. We created a mapping to the SWeMLS Ontology\footnote{We use Ontotext Refine \url{https://www.ontotext.com/products/ontotext-refine/}} and transformed the SMS data into RDF format (\textbf{Step 3A} of \Cref{fig:swemls-process}). 
The generated RDF graphs from Step 3A already connect each system with its respective system components and I/O variables (cf. green arrows on \Cref{figure:example_instance_KG}). However, the connection between I/O variables and components is not yet available (cf. red arrows on \Cref{figure:example_instance_KG}). 
To build these connections, we ran an enrichment process (\textbf{Step 3B}) using SHACL-AF rules\footnote{Example SHACL-AF rules and SHACL validation constraints can be accessed in our GitHub repo, e.g., \url{https://bit.ly/sweml-t3-pattern} for pattern T-3.}. 

Lastly, we validated the resulting data against SHACL constraints (\textbf{Step 3C}). We defined a set of constraints for general SWeMLS as well as instances of specific SWEMLS patterns to ensure completeness, validity and conformance of the KG to the pattern definitions. 

\textit{SWeML-KG update mechanisms.} 
To promote community-based contributions, we published all the source code necessary for the KG creation (i.e., Ontotext refine projects and mapping files, SHACL-AF rules, and SHACL constraints).
Furthermore, we plan to make updating the SWeMLS-KG with new system descriptions easier, for example, by relying on features provided by the Open Research Knowledge Graph to enable community-wide contributions. 
%

\section{Availability of the SWeMLS-KG}
\label{sec:availability}

The SWeMLS-KG landing page\footnote{\url{https://w3id.org/semsys/sites/swemls-kg/}} provides pointers to the various resources covered in this paper, i.e., 
the Linked Data resources\footnote{e.g., Garcia et al. \cite{Garcia2019} \url{https://semantic-systems.net/swemls/System_4QP5XAGX}}, 
the SPARQL query interface\footnote{\url{https://semantic-systems.net/sparql/}}, 
a Zenodo link for the complete RDF snapshots\footnote{\url{https://doi.org/10.5281/zenodo.7445917}}, the source code for SWeMLS toolkit\footnote{\url{https://github.com/semanticsystems/swemls-toolkit}}.
This allows users to choose the most appropriate resources and access mechanisms most suitable for their context.

\paragraph{Publication as ORKG Observatory.} SWeMLS-KG was also made available as part of the Open Research Knowledge Graph~\cite{Auer+2020,10.1145/3360901.3364435}\footnote{\url{https://orkg.org}}. ORKG is a scholarly knowledge organization facility, where contributions conveyed in scientific articles are represented semantically in machine- and human-readable ways.
The FAIR semantic description of research contributions facilitates a number of applications, such as overviews of the state-of-the-art for certain research questions (comparisons), visualizations, or leaderboards.
The ORKG organizes the semantic contribution descriptions along research fields but also in thematic \textit{observatories}, where a team of curators from one or several organizations curates the contributions related to a specific topic.

Together with the ORKG development team, we imported the SWeMLS-KG, so that its data is browsable, accessible, citable and reusable.
The ORKG observatory for SWeMLS-KG allows browsing patterns, instantiations of the patterns as well as searching and filtering contributions from articles by certain characteristics.
An initial version of the SWeMLS-specific ORKG observatory is available online on the ORKG server\footnote{\url{https://orkg.org/observatory/Neurosymbolic_artificial_intelligence}}, providing an overview of the collected papers as well as detailed metadata for each paper\footnote{e.g., Garcia et al. \cite{Garcia2019} in ORKG: \url{https://orkg.org/paper/R574440}}. 

\section{Use Cases}
\label{sec:reusability}

We hereby report on use cases that explore SWeMLS-KG via (i) SPARQL queries (\Cref{sec:querying}, and (ii) Knowledge Graph Embedding methods (\Cref{sec:embedding}). 

\subsection{Use Case 1: Understanding SWeMLS trends through Querying the SWeMLS-KG}
\label{sec:querying}
The SWeMLS-KG can support researchers and reviewers in exploring and understanding trends in the SWeMLS field through queries executed on the SPARQL enpoint of the KG. We first motivate exemplary knowledge questions in natural language, show their SPARQL representations, and discuss their results.

\vspace{1mm}
\begin{lstlisting}[
captionpos=b, 
caption={Query for components in the medical domain for diagnosis prediction}, 
label=lst:componentsQuery,
basicstyle=\scriptsize\ttfamily]
PREFIX swemls: <https://w3id.org/semsys/ns/swemls#>
PREFIX res: <http://semantic-systems.net/swemls/>
/* ... rdf, rdfs, and dc-terms are ommitted*/
select ?swModel ?statisticalModel ?trainingType ?title ?year
where {
  ?system a swemls:System ;
    swemls:hasApplicationDomain res:Domain.Medicine_Health ;
    swemls:hasTask res:Task.Patient_Diagnosis_Prediction ;
    swemls:hasTrainingType / rdfs:label ?trainingType .
  ?system swemls:hasSymbolIO / rdfs:label ?swModel .
  { 
    select ?system (group_concat(?statisticalModelName;separator=",") as ?statisticalModel) 
    where { ?system swemls:hasStatisticalModel / rdfs:label ?statisticalModelName}
    group by ?system 
  }
  ?paper swemls:reports ?system ; terms:title ?title ; swemls:year ?year . 
}
\end{lstlisting}
\vspace{-3mm}

\paragraph{Task/domain driven queries for components of SWeMLS.} We want to support researchers and reviewers in identifying and exploring existing SWeMLS and their components: \textit{What SWeMLS components (SW resources and ML models) have been used to solve a specific task x in the domain y?} A researcher might ask this question e.g., in the course of designing a new system as part of state-of-the-art research or when looking for additional datasets that have been used in a target domain. A reviewer on the other hand, could quickly identify publications with similar components and use these to highlight the innovation and advantages of the submission under review.
A SPARQL representation of this question in the domain \texttt{Medicine\_Health} and for the  task \texttt{Patient\_Diagnosis\_Prediction} is given in Listing \ref{lst:componentsQuery}. \Cref{tab:modelQueryResults} shows an excerpt of the query results, which could be further explored.
%
\begin{table}[htbp]
 \resultsTableFontSize
 \centering
 \setlength\tabcolsep{2pt} 
\vspace{-4mm}
\begin{tabularx}{\textwidth}{l X l X l}
    \textbf{swModel} &
    \textbf{statisticalModel} &
  	\textbf{trainingType} &
    \textbf{title} &
    \textbf{year} 
  \\
\hline

CCS & Attention,GloVe,MLP,RNN & Self-supervised&GRAM: Graph-Based Attention Model for Healthcare Representation Learning&2017\\
UMLS & ARM & Self-supervised & Guiding supervised learning by bio-ontologies in medical data analysis & 2018\\
ICD & Graph-based Attention Model,Knowledge Attention,Gated Recurrent Unit (GRU) & Supervised & KAME: Knowledge-based attention model for diagnosis prediction in healthcare & 2018\\
DBpedia & SVM & Supervised&Improving rare disease classification using imperfect knowledge graph&2019\\
    \hline

\end{tabularx}
\caption{Query results for components in the medical domain for diagnosis prediction (excerpt)}
\label{tab:modelQueryResults}
\vspace{-10mm}
\end{table}

\paragraph{Pattern-driven queries for SWeMLS} are queries that explore the system workflow patterns' structure and its components. This allows researchers and reviewers to identify \textit{structurally} identical or similar SWeMLS  and their relevant aspects, i.e., the integration of ML models and SW resources: \textit{What SWeMLS exist that use a specific SW resource x as input for a ML Model y that produces symbolic output?}

\vspace{1mm}
\begin{lstlisting}[
captionpos=b, 
caption={SPARQL query for systems using a translation model on Facebook benchmark data, producing symbolic output as part of their architecture}, 
label=lst:patternQuery,
basicstyle=\scriptsize\ttfamily]
PREFIX swemls: <https://w3id.org/semsys/ns/swemls#>
PREFIX res: <http://semantic-systems.net/swemls/>
/* ... rdf, rdfs, skos, and dc-terms are ommitted */
select ?domain ?task ?pattern ?sw ?groupSw ?title ?year
where {
  ?system a swemls:System ; swemls:hasApplicationDomain ?domain ; 
    swemls:hasTask ?task ; swemls:hasCorrespondingPattern ?pattern ; swemls:hasStepML ?ml .
  ?paper swemls:reports ?system ; terms:title ?title ; swemls:year ?year .
  ?sw a swemls:SemanticWebResource .
  { ?sw skos:broader res:Resource.Facebook } 
  UNION { 
    select ?sw (group_concat(?compoundSwInput;separator=",") as ?groupSw) 
    where {
      ?sw swemls:hasCompoundElement ?compoundSwInput .
      ?compoundSwInput skos:broader res:Resource.Facebook 
    } group by ?sw 
  }
  ?ml swemls:componentInput ?sw .
  ?ml swemls:componentOutput/rdf:type swemls:SemanticWebResource .
  { ?ml swemls:componentModel res:StatisticalModel.TransX } 
  UNION { 
    ?ml swemls:componentModel ?compoundML .
    ?compoundML swemls:hasCompoundElement res:StatisticalModel.TransX 
  } 
}
\end{lstlisting}
\vspace{1mm}
A SPARQL representation for this question is given in Listing \ref{lst:patternQuery}. We search for systems that use a translation model (TransX) as ML module which operates on Facebook benchmark semantic web resources (e.g., FB15k, FB13, FB500k). Furthermore, the module has to generate symbolic output. The ML module can be placed anywhere in the system architecture, hence, we do not look for specific patterns or architectures.
%
\begin{table}[htbp]
 \resultsTableFontSize
 \centering
 \setlength\tabcolsep{2pt}
\begin{tabularx}{\textwidth}{l X l l l X l}
    \textbf{domain} &
    \textbf{task} &
    \textbf{pattern} &
    \textbf{sw} &
    \textbf{groupSw} &
    \textbf{title} &
    \textbf{year}
  \\
\hline
General & KG\_Completion & F4 & ...SW\_cc6bef6e & FB122 & Jointly embedding knowledge graphs ... & 2016\\
General & KG\_Completion & F2 & ...SW\_d5ee1a61 & FB\_500K & Probabilistic Belief Embedding ... & 2016\\
General & KG\_Completion & A1 & ...SW\_f27afb1c & FB13,FB15k & Learning Knowledge Embeddings by ... & 2017 \\
General & KG\_Completion & A1 & FB15k & & Knowledge Graph Embedding via ... & 2018 \\
General & Question\_Answering & F3 & ...SW\_1fb71cdc & FB15k & Representation Learning of ... & 2019 \\
    \hline

\end{tabularx}
\caption{Query results for systems processing Facebook resources with a translation model, producing symbolic output (excerpt)}
\label{tab:patternQueryResults}
\vspace{-7mm}
\end{table}

%
\Cref{tab:patternQueryResults} shows an excerpt of the found systems, including different patterns and specific SW resources. As an example, the system presented in the paper ``Learning Knowledge Embeddings by Combining Limit-Based Scoring Loss", uses pattern A1 and two Facebook benchmark datasets, namely FB13 and FB15k, for KG completion.

\subsection{Use Case 2: Embedding-based Exploration of the SWeMLS KG}
\label{sec:embedding}
In order to allow further exploration of the SWeMLS-KG, we have computed RDF2vec embeddings~\cite{ristoski2016rdf2vec} on the graph. With the help of those embeddings, visualizations can be generated, and additional queries, based on entity similarity in the embedding space, can be performed. Unlike the previous use case where the exploration of the data is guided by explicitly stated information needs, this use case explores the latent semantics encoded in the SWeMLS-KG.

\begin{figure}[t]
    \centering
    \begin{subfigure}[b]{0.495\textwidth}
        \includegraphics[width=\textwidth]{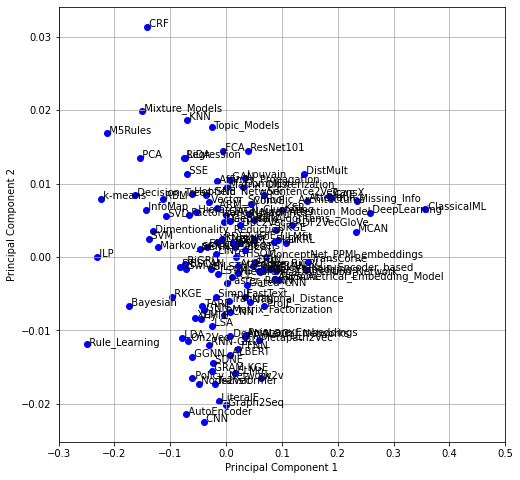}
    \end{subfigure}
    \begin{subfigure}[b]{0.495\textwidth}
        \includegraphics[width=\textwidth]{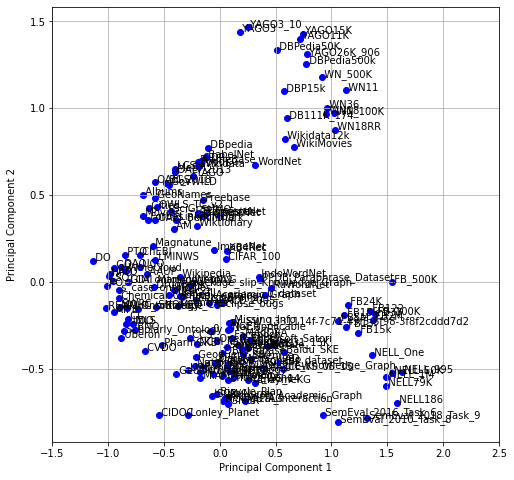}
    \end{subfigure}
    \caption{Scatterplots of statistical model methods (left) and resources (right). For the plots, the embedding spaces have been reduced to 2D using principal component analysis.}
    \label{fig:embedding_scatter}
    \vspace{-4mm}
\end{figure}

Fig.~\ref{fig:embedding_scatter} shows scatter plots of embeddings for the statistical modeling methods and semantic web resources in the knowledge graph. Especially for the resources, one can observe that the grouping is actually sensible, forming, e.g., a cluster of DBpedia and YAGO-related resources in the top area.

Typical scenarios for searching in the embedding space would be triggered using one entity and then searching for further entities in the neighborhood. One typical example use case would be searching for alternative resources and methods. For example, a neighborhood query for the FB15k link prediction benchmark provides a list of other link prediction benchmark datasets, whereas a neighborhood query for DBpedia provides a list of other general-purpose knowledge graphs, such as YAGO or Wikidata. Likewise, neighborhood queries for methods can be conducted. For example, a neighborhood search for graph neural networks gives rule learning and On2vec as nearest neighbors\footnote{Source code:~\url{https://w3id.org/semsys/sites/swemls-kg/rdf2vec}}.

Neighborhood queries can also be useful for finding related papers. We probed the embedding space with a randomly selected paper, describing a knowledge-graph-based recommender system using embeddings, attention networks, and freebase as a resource. The neighborhood contains mostly other papers describing knowledge-based recommenders and papers using the same pattern and/or resources for other purposes, such as question answering.

\section{Related Work}
\label{sec:relatedwork}

\paragraph{Ontologies for describing neuro-symbolic/ML systems.} For the proposed SWeMLS ontology, related work is represented by earlier efforts to characterize neuro-symbolic systems. For example,  Bader and Hitzler \cite{bader2005dimensions} made an early attempt at such characterization and proposed eight dimensions for classification purposes.
More recently, Van Harmelen and ten Teije \cite{VanHarmelen2019a} introduced a set of 13 design patterns, similar to design patterns in software engineering. 
This taxonomy has been extended with processes and models in \cite{VanBekkum2021}.
Another taxonomy comprising six different types of systems but without focus on the internal architectures of the investigated systems has been presented by Kautz \cite{kautzthird}.
While all these efforts focus on the broader family of neuro-symbolic systems, in our recent work~\cite{breit2022combining} we proposed a classification system tailored for SWeMLS. With the ontology presented  here we provide the first machine-actionable (i.e., formally represented) system classification. 

To ensure that ML research outcomes are properly comparable, understandable, reusable, and reproducible several ontologies have been proposed. Onto-DM~\cite{ontodm2008} for instance provides generic representations of entities in data mining, DMOP~\cite{dmop2015} supports meta-learning from ML processes, Exposé~\cite{vanschoren2010expose} can be used to describe and reason about ML experiments, and the MEX Vocabulary~\cite{esteves2015mex} aims to support managing ML outcomes and sharing of provenance data. 
In order to offer a flexible approach for mapping existing ML ontologies and to support extensions, a W3C Community Group\footnote{\url{https://www.w3.org/community/ml-schema/}} developed ML-Schema (MLS)\footnote{\url{http://ml-schema.github.io/documentation/ML Schema.html}}.
Compared to our approach, existing ontologies focus on ML experiment \textit{executions} and not on system descriptions. They also do not focus on representing SW elements of the systems. However, main ML concepts in our ontology can be mapped to ML-Schema, such as \texttt{mls:Experiment}, which is comparable to \texttt{swemls:System}, and \texttt{mls:Data} is similar to \texttt{swemls:Instance}.
With a focus on reproducibility, ML-Schema and other approaches also cover detailed ML settings, such as hyperparameters and evaluation results. While this is not in the focus of our current work, we plan to extend our knowledge graph in this direction. 

\paragraph{Machine-processable publication of domain-specific scientific knowledge} is reported in several disciplines. For example, in social sciences, in the domain of human cooperation, experts annotated nearly 3,000 studies in terms of 60 features as part of the COoperation DAtabank initiative\footnote{COoperation DAtabank: \url{https://amsterdamcooperationlab.com/databank/}}. This systematically collected scientific knowledge has been published using semantic technologies as a knowledge graph~\cite{TiddiESWC2020} and as nanopublications~\cite{ImranEScience21} in order to support the automation of scientific tasks such as (comparative) meta-analysis and the detection of contradictory claims respectively.  

Going beyond domain-specific efforts for publishing scientific knowledge, the Open Research Knowledge Graph (ORKG) aims to provide a platform for the publication of open research knowledge. ORKG describes scientific articles semantically in a machine- and human-readable way. It offers concepts and properties to classify articles, and extract various metadata such as authors and publication date, but also to describe research contributions and results. To contribute to this initiative, we mapped our ontology to the ORKG schema and published our SWeMLS results within the ORKG (see \Cref{sec:availability}), thus being one of the first communities to leverage the capabilities of this system and to benefit by the sustainability of the data publication on a long term.

\section{Conclusion and Future Work}
\label{sec:conclusion}
%
In this work, we used a semantic technology approach to provide a machine-processable way to represent a large number of SWeMLS reported in scientific publications. We introduced the SWeMLS ontology and SWeMLS knowledge graph to support researchers and reviewers using a more automated approach to search, conduct analysis and test existing SWeMLS. The SWeMLS-KG was also imported into  ORKG, making the data browseable, accessible, citable, and reusable. The use cases we discussed have shown that the SWeMLS-KG is useful for researchers and reviewers on a variety of levels, including identifying and analyzing existing SWeMLS, drawing conclusions about the components being used, or identifying similar components using SPARQL queries and embedding-based exploration of the SWeMLS-KG.

Regarding future work, we plan to include audit support for SWeMLS by capturing Workflow Execution traces to complement the workflow templates (i.e., SWeMLS patterns) and workflow instances (i.e., SWeMLS instance), building on our prior work on auditability~\cite{ekaputra2021semantic}.
Furthermore, we strive to enable semi-automatized description extraction from SWeMLS papers and generation of SWeMLS pipeline code from patterns by building on existing works~\cite{daga2023data,grafberger2022data}.
Finally, we want to support a two-way transformation of data from the SWeMLS-KG and the ORKG-observatory.
Beyond the scope of our research, we hope to inspire broader research communities to provide their research results in a structured representation, which in turn will allow others to build their research \textit{by standing on the shoulder of giants}.

\section*{Acknowledgments}
\label{ack}
This work has been supported by the Austrian Science Fund (FWF) under grant V0745 (HOnEst) and FFG Project OBARIS (Grant Agreement No 877389).
SBA Research (SBA-K1) is a COMET Center within the COMET – Competence Centers for Excellent Technologies Programme and funded by BMK, BMAW, and the federal state of Vienna. The COMET Programme is managed by FFG.
Moreover, financial support by the Christian Doppler Research Association, the Austrian Federal Ministry for Digital and Economic Affairs, the National Foundation for Research, Technology and Development, DFG NFDI4DataScience (No. 460234259) and ERC ScienceGRAPH (GA ID: 819536) is gratefully acknowledged.

\vskip 1cm

\bibliographystyle{bib/splncs04}
\bibliography{bib/bibliography.bib,bib/standards.bib}

\end{document}